\documentclass[letterpaper, 10 pt, conference]{ieeeconf}  

\IEEEoverridecommandlockouts                              

\makeatletter
\let\NAT@parse\undefined
\makeatother
\usepackage[numbers,sort&compress]{natbib}

\usepackage[pdftex]{graphicx}
\usepackage{amsmath}
\usepackage{amssymb}  
\usepackage{subfigure}
\usepackage{subcaption}
\usepackage{multirow}
\usepackage{array,booktabs}
\usepackage{diagbox}
\usepackage{balance}
\usepackage{xspace}
\usepackage{algorithm}
\usepackage{algpseudocode}
\usepackage{adjustbox}
\usepackage{romannum}

\usepackage[final]{hyperref}
\hypersetup{
 colorlinks=true,
 linkcolor=blue,
 filecolor=magenta,
 urlcolor=blue,
 citecolor=black
}
\usepackage{cleveref}
\crefname{figure}{Fig.}{Figs.}
\Crefname{figure}{Fig.}{Figs.}
\usepackage[font=small]{caption}
\usepackage{rpm_SIunits}
\usepackage{rpm_acronyms}
\usepackage{rpm_math}
\usepackage{rpm_misc}

\usepackage{soul,color}
\usepackage{lipsum}

\usepackage{xcolor}

\usepackage{tikz} 
\usepackage{subcaption}
\usepackage{caption}
\usepackage[font=scriptsize]{caption}
\title{\LARGE \bf GaRLIO: Gravity enhanced Radar-LiDAR-Inertial Odometry
}     

\author{Chiyun Noh${}^{1}$, Wooseong Yang${}^{1}$, Minwoo Jung${}^{1}$, Sangwoo Jung${}^{1}$ and Ayoung Kim${}^{1*}$
\thanks{$^\dagger$This work was supported by the National Research Foundation of Korea(NRF) grant funded by the Korea government(MSIT) (No. RS-2023-00241758), and in part by the Robotics and AI (RAI) Institute.}
\thanks{$^{1}$C. Noh, W. Yang, M. Jung, S. Jung and A. Kim are with the Dept. of Mechanical Engineering, SNU, Seoul, S. Korea {\tt\small [gch06208, yellowish, moonshot, dan0130, ayoungk]@snu.ac.kr}}%
}
\begin{document}

\maketitle
\thispagestyle{empty}
\pagestyle{empty}

\begin{abstract}
Recently, gravity has been highlighted as a crucial constraint for state estimation to alleviate potential vertical drift. 
Existing online gravity estimation methods rely on pose estimation combined with IMU measurements, which is considered best practice when direct velocity measurements are unavailable. However, with radar sensors providing direct velocity data—a measurement not yet utilized for gravity estimation—we found a significant opportunity to improve gravity estimation accuracy substantially. GaRLIO, the proposed gravity-enhanced Radar-LiDAR-Inertial Odometry, can robustly predict gravity to reduce vertical drift while simultaneously enhancing state estimation performance using pointwise velocity measurements. Furthermore, GaRLIO ensures robustness in dynamic environments by utilizing radar to remove dynamic objects from LiDAR point clouds. Our method is validated through experiments in various environments prone to vertical drift, demonstrating superior performance compared to traditional LiDAR-Inertial Odometry methods. We make our source code publicly available to encourage further research and development.
\url{https://github.com/ChiyunNoh/GaRLIO}
\end{abstract}

\section{Introduction}
\label{sec:intro}

Range-based sensor \ac{SLAM} is widely used for precise pose estimation of mobile robots in unknown environments. Along with many of range sensors, \ac{LiDAR} has drawn attention for its intuitive spatial information, especially being integrated with \ac{IMU} to exhibit high precision and robustness \cite{FAST-LIO, LINS, POINT-LIO}. However, due to the sparse vertical resolution and remaining uncertainties in \ac{LiDAR} measurements \cite{lidar_bias}, \ac{LIO} is still susceptible to vertical drift. 

In this context, exploiting gravity shows promising performance in mitigating vertical drift from recent works \cite{kubelka2022gravity, D-LIOM, nemiroff2023joint,wildcat,agha2021nebula}. Since gravity maintains size and direction constant across diverse conditions, precise local gravity estimation allows for an accurate rotation calculation between the global and body frame. To be more specific, accurate gravity estimation may lead to reliable observation of roll and pitch, reducing the vertical position error. Most existing methods measure gravity vector relying on the relationship between \ac{IMU} acceleration measurements and estimated pose \cite{D-LIOM, nemiroff2023joint,wildcat} or only when the robot motion is stationary \cite{agha2021nebula}. However, establishing the relationship between pose and acceleration generally results in amplified errors due to the double integrated bias and noise of the \ac{IMU} acceleration \cite{forster2016manifold}.

Since traditional \ac{LIO} lacks direct velocity observations at the raw measurement level, they rely on pose estimation and \ac{IMU} measurements to estimate gravity. Yet, we observe that incorporating velocity measurements can significantly enhance the accuracy of gravity estimation. While one study has explored the fusion of radar with \ac{LIO} to improve state estimation performance by providing velocity measurements \cite{DR-LRIO}, no existing work has integrated velocity measurements specifically for the purpose of gravity estimation.

\begin{figure}[!t]
    \centering
    \includegraphics[width=\linewidth]{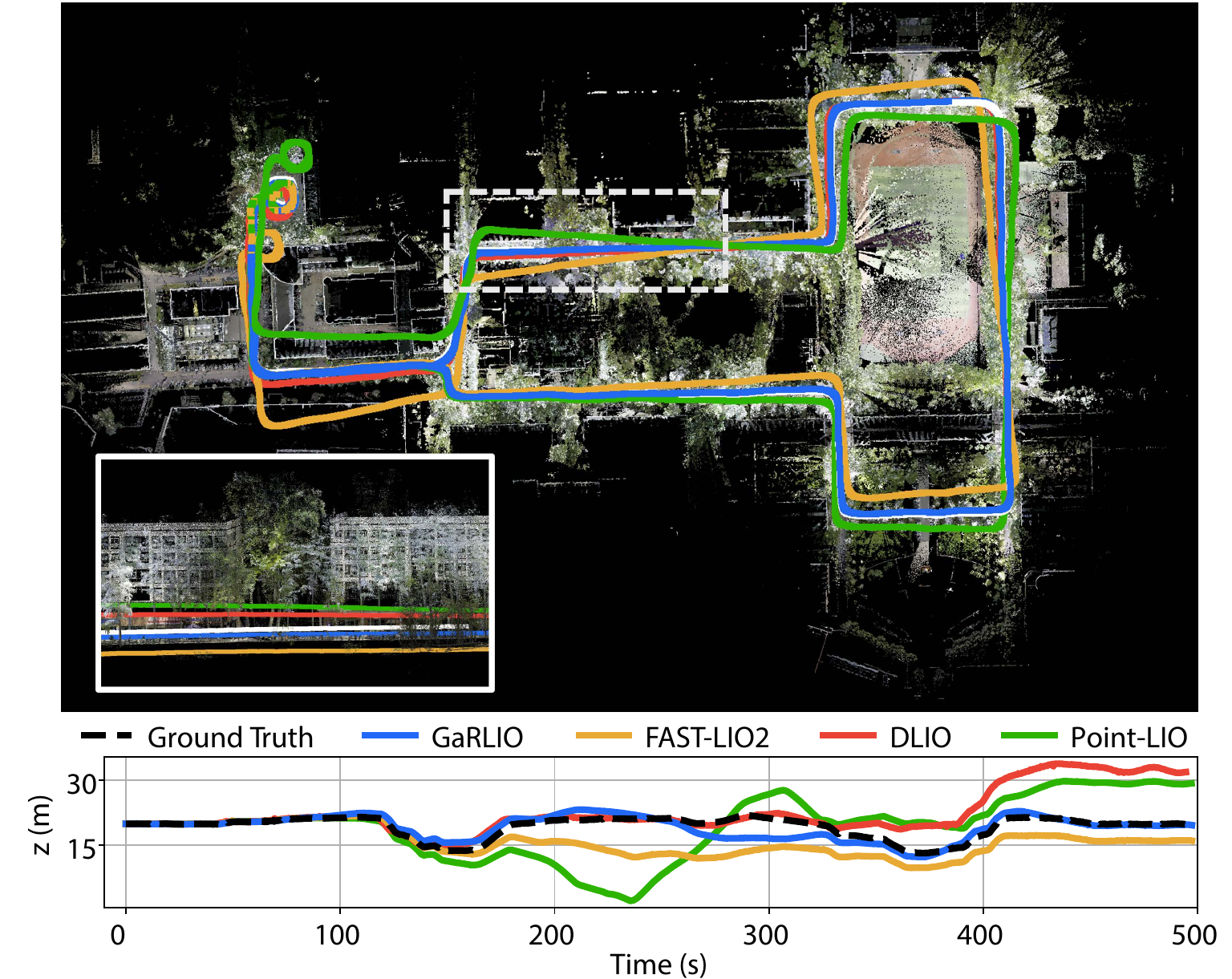}
    \caption{\textbf{Top}: Trajectories of GaRLIO and other methods with ground truth (white) overlaid on the \ac{TLS} map. \textbf{Bottom}: Elevation plot along path length. Our method (blue) reported only $\unit{1.21}{m}$ vertical mean error over $\unit{2.045}{km}$ path length.}
    \label{fig:pipeline}
    \vspace{-7mm}
\end{figure}

\begin{figure*}[t!]
    \centering
    \includegraphics[width=0.8\textwidth]
    {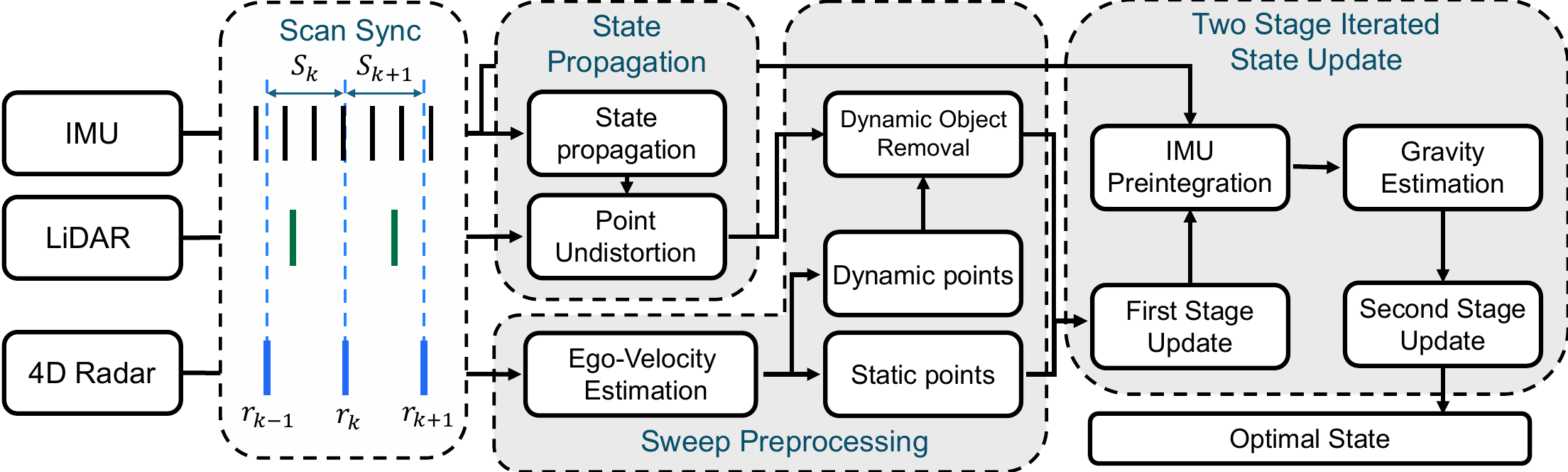}
    \caption{
    GaRLIO is divided into four primary modules. Each module contributes to achieving the optimal state by removing \ac{LiDAR} dynamic points and calculating both pointwise velocity and velocity-aware gravity residuals.
    }
    \label{fig:overview}
    \vspace{-8mm}
\end{figure*}

In this paper, we propose GaRLIO, a gravity-enhanced Radar-\ac{LiDAR}-Inertial Odometry that introduces a novel idea of exploiting radar for gravity estimation. GaRLIO integrates a new gravity estimation technique that leverages radar Doppler measurements to address vertical drift and inaccuracy. Integrating radar with \ac{LIO} improves velocity estimation accuracy, allowing for more precise local gravity estimation, which reduces vertical drift. Furthermore, this integration facilitates the dynamic object filtering in \ac{LiDAR} point cloud, ensuring performance in diverse environments. GaRLIO is evaluated on public datasets with various challenging scenarios, including downhill conditions where vertical drift prevalently occurs in \ac{LIO}, exhibiting surpassing performance compared with \ac{SOTA} methods. 
The pipeline of our method is as shown in \figref{fig:overview}. Our contributions are as follows:
\begin{itemize}
    \item We propose GaRLIO that utilizes gravity to address the vertical drift of \ac{LIO}. GaRLIO introduces a novel velocity-aware gravity estimation that leverages radar Doppler measurements to reduce vertical drift and inaccuracy common in velocity-ignorant approaches. To our knowledge, this is the first method using radar velocity to estimate gravity.
    \item GaRLIO fuses radar with \ac{LIO} and demonstrated enhanced state estimation performance by exploiting pointwise velocity residuals based on radar measurements. Additionally, our fusion approach effectively removes dynamic points within \ac{LiDAR} point clouds. 
    \item GaRLIO is evaluated on public datasets with challenging scenarios, including downhill, where the introduced residuals demonstrated performance improvement. We open-source GaRLIO to encourage further development in radar-\ac{LiDAR}-\ac{IMU} fusion.
\end{itemize}

\section{related work}
\label{sec:relatedwork}

\subsection{LiDAR and Radar Fusion}

Radar and \ac{LiDAR} are complementary range sensors, prompting numerous studies on their fusion for localization \cite{ROLM, YeongSang_radar, RALL}. \citeauthor{ROLM}~\cite{ROLM} and \citeauthor{YeongSang_radar}~\cite{YeongSang_radar} leveraged a \ac{LiDAR} map to align radar scans, enabling robust localization in adverse conditions as smoke or fog. However, they heavily depend on an accurate \ac{LiDAR} map, underscoring the importance of combining both sensors for precise \ac{SLAM}.

Despite the significance of sensor fusion, \ac{LiDAR} and radar fusion for \ac{SLAM} remains uncommon. Even though DR-LRIO \cite{DR-LRIO} proposed a tightly-coupled method between \ac{FMCW} radar and \ac{LiDAR} to enhance localization in challenging environments, it still contends with inherent uncertainties of radar, particularly along the $z$-axis. Our method expands the application of radar data beyond ego-velocity estimation by introducing additional strategies, such as dynamic object filtering and gravity estimation, to further improve \ac{SLAM} performance.

\subsection{Gravity Estimation in LiDAR-Inertial Odometry}
Accurate gravity estimation is essential in \ac{LIO} since \ac{IMU} acceleration measurements include gravity. Additionally, aligning maps based on the gravity vector alleviates vertical drift. The most common approach for predicting gravity involves probabilistic methods \cite{FAST-LIO, LINS}. However, these studies did not focus on exploiting gravity for state updates or map alignment. Recent works directly impose gravity as new constraints to mitigate vertical drift. Nebula \cite{agha2021nebula} measured gravity using \ac{IMU} acceleration only when the robot is stationary, introducing a gravity factor to constrain roll and pitch. D-LIOM \cite{D-LIOM} and Wildcat \cite{wildcat} estimated gravity using \ac{IMU} measurements and odometry, incorporating gravity alignment constraints. Additionally, \citeauthor{nemiroff2023joint} \cite{nemiroff2023joint} jointly optimized accelerometer intrinsics and gravity.
These methods partially tackled vertical drift; however, relying on velocity-ignorant models for gravity estimations. The velocity-ignorant model focuses on the relationship between \ac{IMU} acceleration and the pose, which impedes accurate gravity estimation due to errors exacerbated by double integration. To tackle these issues, we incorporate radar-based velocity residuals and velocity-aware gravity residuals into \ac{LIO}, eliminating the necessity of double integration.

\section{Preliminary}
\label{sec:method}

\subsection{Notation}
\label{subsec: Notation}
In this paper, $x_k^j$ denotes the $j_{th}$ iteration of Kalman filter update for $k_{th}$ state, and $(), \widehat{()}, \bar{()}$ denote ground truth, propagated, and optimal state. Global frame ${}^G()$ is set as initial \ac{IMU} frame, while ${}^A()$ denote sensor frames. The state description and definition of $\boxplus/\boxminus$ are as follows:

\vspace{-4mm}\small
\begin{eqnarray}
\label{eq:state}
	\mathcal{M} &\triangleq& SE_{2}(3) \times \mathbb{R}^{6}, \:   
	    \textbf{x} \: \triangleq \: [\mathcal{X}^T, \: \textbf{b}_{\omega}^T, \: \textbf{b}_a^T ]^T \in\mathcal{M} \nonumber  \\
	\textbf{u} &\triangleq& \left[\omega_m^T \quad  a_m^T \right]^T,\, \: \textbf{n} \: \triangleq \: \left[n_\omega^T, \: n_a^T, \: n_{b\omega}^T, \: n_{ba}^T  \right]^T \\
    \label{eq:SE2(3) define}
    \mathcal{X}  &\triangleq& \begin{bmatrix}
    {}^G\textbf{R}_{I} & {}^G\textbf{v}_{I} & {}^G\textbf{p}_{I} \\
     0 & 1 & 0 \\
     0 & 0 & 1 \\
    \end{bmatrix}\in SE_2(3) \nonumber  \\
     \delta \widehat{\textbf{x}}_{k}&\triangleq&\textbf{x}_{k} \boxminus \widehat{\textbf{x}}_{k} = \left [ \delta \widehat{\theta}_{k}, \delta \widehat{\textbf{v}}_{k},\delta \widehat{\textbf{p}}_{k},\delta \widehat{\textbf{b}}_{\omega_{k}},\delta \widehat{\textbf{b}}_{a_k} \right ]  \nonumber\\
     &=& \left [\mathrm{Log}( \mathcal{X}_k\widehat{\mathcal{X}}_k^{-1}), {\textbf{b}}_{\omega_{k}}-\widehat{\textbf{b}}_{\omega_{k}}, {\textbf{b}}_{a_k}- \widehat{\textbf{b}}_{a_k} \right ] \in \mathbb{R}^{15}
     \nonumber \\
     \textbf{x} \boxplus \delta \widehat{\textbf{x}} &=& \begin{bmatrix}\mathcal {X} \\ \mathbf {b} \end{bmatrix} \boxplus \begin{bmatrix} {\xi } \\ \widehat{\mathbf {b}} \end{bmatrix} \triangleq \begin{bmatrix}\rm Exp({\xi }) \cdot \mathcal {X} \\ \mathbf {b} + \widehat{\mathbf {b}} \end{bmatrix} \nonumber
     \vspace{-8mm}
\end{eqnarray}
\normalsize
, where $\mathbf {b}, \widehat{\mathbf {b}} \in \mathbb{R}^{6}$ and ${\xi} \in \mathbb{R}^{9}$. In \eqref{eq:state}, ${}^G\textbf{R}_{I}$, ${}^G\textbf{v}_{I}$, and ${}^G\textbf{p}_{I}$ denote the rotation, velocity, and position of the \ac{IMU} in the global frame, while $\textbf{b}$ represents the IMU bias. The angular velocity and linear acceleration from the \ac{IMU} are denoted by $\omega_m$ and $a_m$, with $\textbf{n}$ representing the Gaussian white noise associated with these measurements and biases.

\subsection{State Propagation}
\label{subsec: State Propagation}
For state prediction within the time interval $[k, k+1)$, we discretize the continuous-time \ac{IMU} kinematic model using a zero-order holder with an \ac{IMU} sampling period $\Delta t$ \cite{FAST-LIO}. The resulting discretized motion model \( \textbf{f} \), is formulated as:

\vspace{-5mm}\small
\begin{align}
\label{eq:propagate}
	\textbf{x}_{i+1} &= \textbf{x}_{i} \boxplus \left(\Delta t \, \textbf{f}\left(\textbf{x}_i, \textbf{u}_i, \textbf{n}_i\right)\right), \\
	\textbf{f}\left(\textbf{x}_i, \textbf{u}_i, \textbf{n}_i\right) &= \begin{bmatrix} 
    	\omega_{m_i} - b_{\omega_i} - n_{\omega_i} \\
    	{}^G\textbf{R}_{I_i} \left(a_{m_i} - b_{a_i} - n_{a_i}\right) + {}^G\textbf{g}_i \\
        {{}^G\textbf{v}}_{I_i} + \frac{1}{2}\left({}^G\textbf{R}_{I_i} \left(a_{m_i} - b_{a_i} - n_{a_i}\right) + {}^G\textbf{g}_i\right) \Delta t \nonumber \\
    	n_{b\omega_i} \\
    	n_{ba_i} 
    \end{bmatrix}
\end{align}
\normalsize
State prediction is conducted using the discrete \ac{IMU} kinematic model from \eqref{eq:propagate} by setting the noise term $\textbf{n}=0$:

\vspace{-3mm}\small
\begin{eqnarray}
\label{eq:state_predict}
\widehat{\textbf{x}}_{\tau+1} = \widehat{\textbf{x}}_{\tau} \boxplus \left(\Delta t \textbf{f}\left(\widehat{\textbf{x}}_\tau, \textbf{u}_\tau, 0\right)\right)\:;\;\;t_k \leq t_\tau <t_{k+1} 
\vspace{-1mm}
\end{eqnarray}
\normalsize

\subsection{Error State Prediction}
\label{subsec:Error State Prediction}
To ensure the system depends on the error rather than the state, we define the error using a right-invariant error \cite{hartley2020contact}. The error-state transition matrix is derived by combining the error state \eqref{eq:state} with the \ac{IMU} motion model \eqref{eq:propagate}. The error-state $\delta \widehat{\textbf{x}}_{\tau+1}$ and its covariance $\widehat{\textbf{P}}_{\tau+1}$ are then computed as:

\vspace{-5mm}\small
\begin{align}
\label{eq:nominal state prop}
  \delta \widehat{\textbf{x}}_{\tau+1} =& \: {\textbf{x}}_{\tau+1} - \widehat{\textbf{x}}_{\tau+1}
  \simeq \textbf{F}_{\delta \widehat{\textbf{x}}_{\tau}}\delta \widehat{\textbf{x}}_{\tau} + \textbf{F}_{\textbf{n}_\tau} \textbf{n}_\tau \\ 
  \label{eq:transition matrix}
  \widehat{\textbf{P}}_{\tau+1} =& \: \textbf{F}_{\delta \widehat{\textbf{x}}_{\tau}}\widehat{\textbf{P}}_{\tau}\textbf{F}^T_{\delta \widehat{\textbf{x}}_{\tau}} + \textbf{F}_{\textbf{n}_\tau}\textbf{Q}_{\tau}\textbf{F}^T_{{\textbf{n}}_\tau} \nonumber
\end{align}
\normalsize

The initial value of $\delta \widehat{\textbf{x}}_{i}$ is set to zero. $\textbf{F}_{\delta \widehat{\textbf{x}}_{\tau}}$ and $\textbf{F}_{\textbf{n}_\tau}$ denote the Jacobian matrix of $\delta \widehat{\textbf{x}}_{\tau+1}$ with respect to $\delta \widehat{\textbf{x}}_{\tau}$ evaluated under the condition that $\delta \widehat{\textbf{x}}_{\tau}=0$ and $\textbf{n}_\tau=0$.

\subsection{Iterated State Update}
\label{subsec:Iterated State Update}
The state $\widehat{\textbf{x}}_{k+1}$ and covariance $\widehat{\textbf{P}}_{k+1}$ are used as the prior distribution of $\textbf{x}_{k+1}$. With the measurement at time index $k+1$ serving as the observation, the posterior distribution of the state can be determined \cite{he2021kalman}. Using the prior distribution, the error state propagation can be performed as follows:

\vspace{-3mm}\small
\begin{eqnarray}
\begin{aligned}
  \label{eq:prior distribution}
    \delta\widehat{\textbf{x}}_{k+1}
    &=(\widehat{\textbf{x}}_{k+1}^j\boxminus\widehat{\textbf{x}}_{k+1})+\textbf{J}_{k+1}\delta\textbf{x}_j \\
    \delta\textbf{x}_j&\sim\mathcal{N}(-{\textbf{J}_{k+1}^{-1}}(\widehat{\textbf{x}}_{k+1}^j\boxplus\widehat{\textbf{x}}_{k+1}),{\textbf{J}_{k+1}^{-1}}\widehat{P}_{k+1}{\textbf{J}_{k+1}^{-\top}})
\end{aligned}
\end{eqnarray}
\normalsize
$\textbf{J}_{k+1}$ is the Jacobian of $\delta\widehat{\textbf{x}}_{k+1}$ w.r.t $\delta\textbf{x}_j$ at 0, and detailed derivation is in \cite{shi2023invariant}. After project $\widehat{\textbf{x}}_{k+1}^j$ to the same space of $\delta\textbf{x}_j$, we can get an optimal $\delta\textbf{x}_j$ combining \eqref{eq:prior distribution} and measurement model  $z_{k+1}^M = \textbf{h}_{M^j}(\textbf{x}_k,\textbf{v}_j^M)$ :

\vspace{-3mm}\small
\begin{eqnarray}
\label{eq:MAP}
  \min\limits_{\delta\textbf{x}_j} ( \lVert \textbf{x}_{k+1} \boxminus \widehat{\textbf{x}}_{k+1} \rVert ^2_{\textbf{P}} + \sum_{M}\sum^{|M|}_{j=1} \lVert \textbf{r}_{M^j} +\textbf{H}_{M^j}\delta \widehat{\textbf{x}}_k \rVert ^{2}_{{R_j^M}} )
\end{eqnarray}
\normalsize
where $\textbf{P}={\textbf{J}_{k+1}^{-1}}\widehat{\textbf{P}}_{k+1}{\textbf{J}_{k+1}^{-\top}}$, $\textbf{H}_{M^j}$ is the Jacobian matrix of ${\textbf{h}}_{M^j}$ w.r.t $\delta \widehat{\textbf{x}}_{k}$ and measurement model which is used in this paper is described in \secref{subsec:first_update} and \secref{subsec:second_update}. The $\delta x_j$ that minimizes equation \eqref{eq:MAP} can be obtained through the Kalman filter update process using $\mathbf{H}=\left[\mathbf{H}_{M^1},\cdots \right], \mathbf{R}=diag\left[{R}_1^M,\cdots\right]$. If the state converges, the optimal state $\bar{\textbf{x}}_{k+1}$ and covariance $\bar{\mathbf{P}}_{k+1}$ are updated and details are in \cite{he2021kalman}.

\section{Gravity-enhanced Radar-LiDAR Fusion}
\subsection{Radar-LiDAR Scan Synchronization}
\label{subsec:Scan_Synchronization}
Since 4D radar operates at a higher frequency than \ac{LiDAR}, we utilized sweep reconstruction \cite{SR-LIVO} to synchronize the end timestamp of \ac{LiDAR} and radar, using the radar's timestamp $r_k$. A sweep $S_{k+1}$ is constructed using \ac{LiDAR} points and \ac{IMU} measurements recorded between $r_k$ and $r_{k+1}$, along with radar captured at $r_{k+1}$. To address possible missing segments in the radar-based sweep reconstruction, we filled in missing \ac{FOV} sections of $S_{k+1}$ with data from the previous sweep $S_k$ to generate a complete \ac{LiDAR} scan. 
Once the sweep $S_{k+1}$ is constructed, state propagation is performed using \ac{IMU} measurements as \eqref{eq:state_predict}. \ac{LiDAR} points in $S_{k+1}$ are subsequently corrected with motion compensation to be transformed into the $\left ( \cdot \right )^{r_{k+1}}$ frame.

\subsection{Radar-guided LiDAR Sweep Preprocessing}
\label{subsec:Scan Filtering}
Before the sweep $S_{k+1}$ is utilized for the update module, a preprocessing step is conducted to remove dynamic points from the sweep based on radar Doppler measurements.
\subsubsection{Radar Point Cloud Pre-processing}
The 4D radar provides both 3D points and Doppler velocities, enabling it to compute 3D ego velocity in the $\left ( \cdot  \right )^{r_k}$ frame through the 3-Point RANSAC-LSQ method \cite{ekf-radar}. With this velocity, radar can filter out moving objects within a single frame without needing sequential data. 
This filtering segments the radar point cloud into static points, ${}^RP_s$, and dynamic points, ${}^RP_d$, which include both moving objects and noise.

\subsubsection{Dynamic Removal in LiDAR}
Dynamic points ${}^RP_d$ are exploited to remove dynamic points from the \ac{LiDAR} point cloud, ${}^LP$. However, due to the higher uncertainty in the z-direction and sparser nature of the radar point cloud used in this work, compared to the \ac{LiDAR} point cloud, establishing precise point correspondences becomes challenging.

To overcome this, we projected both the \ac{LiDAR} and radar dynamic point clouds onto the $xy$ plane and compared the two in 2D space. Using the pointwise uncertainty matrix $\Sigma_{r_{i}}$ \cite{4DRadarSLAM}, we computed the Mahalanobis distance between each \ac{LiDAR} point and the radar dynamic point as follows:

\vspace{-3mm} \small
\begin{eqnarray}
\begin{aligned}
    \label{eq:lidar_dynamic}
    d=(P_{XY}({}^Rp_i-{}^R\textbf{T}_L{}^Lp_j))^\top\bar{\Sigma}_{r_i}(P_{XY}({}^Rp_i-{}^R\textbf{T}_L{}^Lp_j)),
\end{aligned}
\end{eqnarray}
\normalsize
where $P_{XY}$ is the projection matrix on $xy$ plane, $\bar{\Sigma}_{r_i} = P_{XY}\Sigma_{r_i}, {}^Rp_i\in{}^RP_d, {}^Lp_j\in{}^LP$. \ac{LiDAR} points within a threshold $d<\epsilon$ are considered dynamic and filtered out.

\subsection{Pointwise Residuals for First Stage Update}
\label{subsec:first_update}
In the first stage update, radar and \ac{LiDAR} measurements are integrated to update the state using pointwise residuals.

\subsubsection{Point-to-Plane Residual}
The residual is calculated using the dynamic-removed \ac{LiDAR} point cloud, ${}^LP_{k+1}$, obtained from \secref{subsec:Scan Filtering}. Each point ${}^Lp_j$ in ${}^LP_{k+1}$ is first transformed into the global frame. Then, five neighboring points are selected for each point using an ikd-tree \cite{cai2021ikd}, and the residual for each point is computed as follows:

\vspace{-4mm}\small
\begin{eqnarray}
\begin{aligned}
  \label{eq:point_to_plane residual}
    \textbf{h}_{L^j}(\textbf{x}_k,\textbf{v}_j^L)=&\\{}^Gu_j^\top({}^G\textbf{R}&_{I_k}({}^I\textbf{R}_L({}^Lp_j+\textbf{v}_j^L)+{}^Ip_L)+{}^Gp_{I_k}-{}^Gq_j),
\end{aligned}
\end{eqnarray}
\normalsize
where ${}^I\textbf{R}_L$, ${}^Ip_L$ are \ac{LiDAR}-\ac{IMU} extrinsic parameter, ${}^Gq_j$ is the centroid of five points, ${}^Gu_j$ is their normal vector and $\textbf{v}_j^L$ is a measurement noise of \ac{LiDAR}. Based on the residuals in \eqref{eq:point_to_plane residual}, the Jacobian matrix of ${\textbf{h}}_{L^j}$ w.r.t $\delta \widehat{\textbf{x}}_{k}$, $\textbf{H}_{L^j}$ is computed according to the definition of the invariant error: 

\vspace{-4mm}\small
\begin{eqnarray}
  \label{eq:LiDAR Measurement Model}
    \textbf{H}_{L^j}=
    {}^Gu_j^\top\left[-\lfloor{}^Gp_{j}\rfloor_\times \quad \textbf{0}\quad\textbf{I}\quad\textbf{0}\quad\textbf{0}\right] \nonumber\\
    \textbf{r}_{L_j}+\textbf{H}_{L^j}\delta\widehat{\textbf{x}}_k\sim\mathcal{N}(0,R_j^L)
\end{eqnarray}
\normalsize
where ${}^Gp_{j}$ is \ac{LiDAR} point in global frame and we set $R_j^L$ as the same value.

\begin{figure}[!t]
    \centering
    \includegraphics[width=1\linewidth]{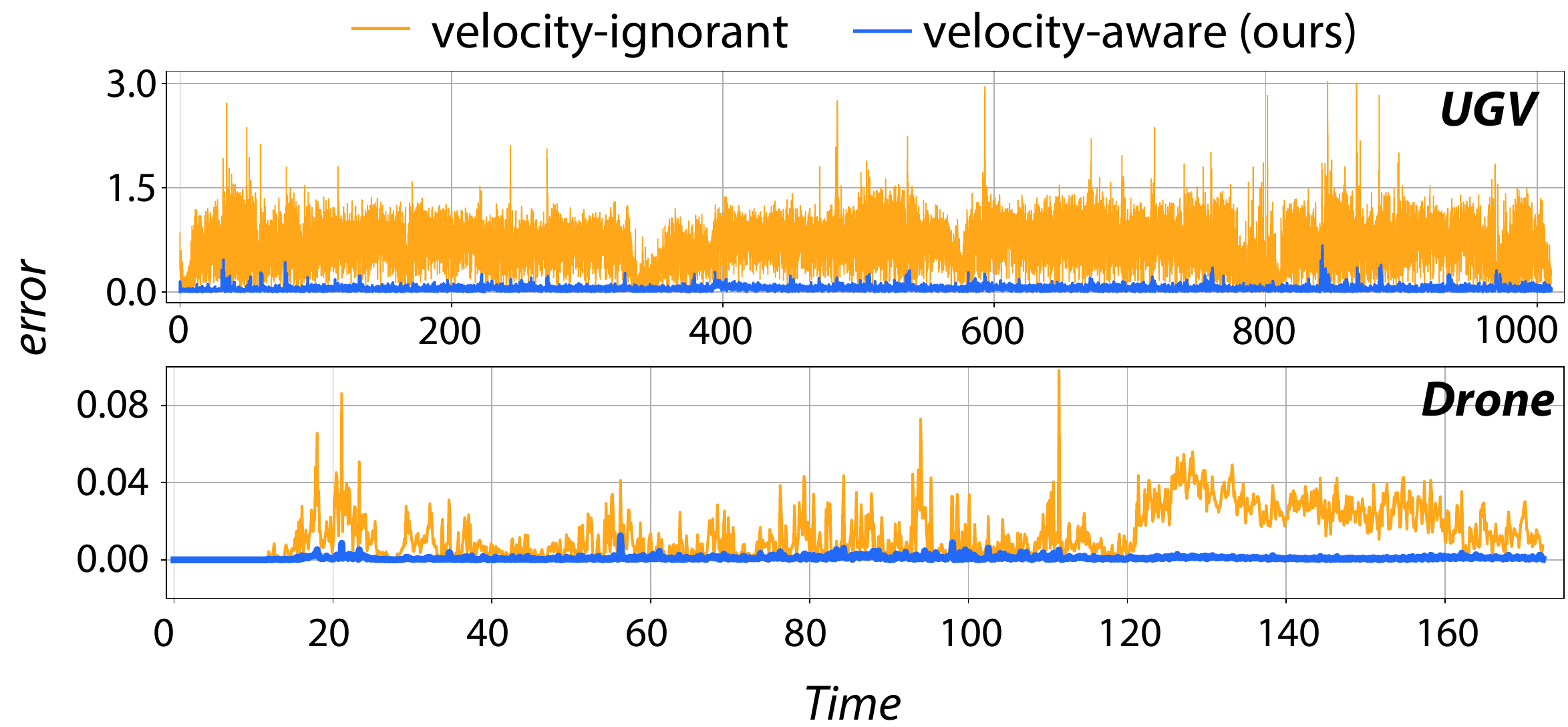}
    \vspace{-6mm}
    \caption{Gravity estimation using the velocity-ignorant (yellow) and velocity-aware method (ours, blue) on different platforms.}

    \label{fig:double integration}
    \vspace{-8mm}
\end{figure}

\subsubsection{Pointwise Velocity Residual}
The static radar points ${}^Rp_j$ are exploited as measurements to compute the velocity residual. The pointwise velocity residual is as follows:

\vspace{-4mm}\small
\begin{align}
    \textbf{h}_{R^j}(\textbf{x}_k,\textbf{v}_j^R)
    &= \textbf{u}({}^Rp_j)^\top {}^G\widehat{\textbf{v}}_R-{}^R\textbf{v}_{m,j} \\
    &=\textbf{u}({}^Rp_j)^\top{}^I\textbf{R}_R^\top\left({}^G\textbf{R}_{I_k}{}^G\textbf{v}_I+\lfloor \omega_I \rfloor_{\times}{}^Ip_R\right)-{}^R\textbf{v}_{m,j} \nonumber
  \label{eq:pointwise velocity residual}
\end{align}
\normalsize
where ${}^I\textbf{R}_R$ and ${}^I\textbf{p}_R$ are the radar-\ac{IMU} extrinsic parameters, and ${}^R\textbf{v}_{m,j}$ is the Doppler measurement at point ${}^R\textbf{p}_j$. The function $\textbf{u}(\cdot)$ represents the unit vector.

Since static radar points may still contain outliers, each static point's uncertainty is assessed using the modified z-score $\mathrm{M_j}$, which is based on deviation between ${}^R\textbf{v}_{m,j}$ and the estimated ego velocity from \secref{subsec:Scan Filtering}. Utilizing these residuals and uncertainties, the Jacobian of the pointwise velocity residual can be computed as follows:

\vspace{-4mm}\small
\begin{eqnarray}
\begin{aligned}
  \label{eq:radar measurement model}
    \textbf{H}_{R^j}&=
    \textbf{u}({}^Rp_j)^\top {}^I\textbf{R}_R^\top\left[\textbf{0}\quad{}^G\widehat{R}_{I_k}^\top\quad\textbf{0}\quad\lfloor{}^Ip_{R}\rfloor_\times\quad\textbf{0}\right] \nonumber\\
    \textbf{r}_{R^j}&+\textbf{H}_{R^j}\delta\widehat{\textbf{x}}_k=-n_{R,j}\sim\mathcal{N}(0,R_j^R) \nonumber
\end{aligned}
\end{eqnarray} \vspace{-3mm}
\begin{equation}
       R_j^R = 
        \begin{cases}
            R_v+\alpha\mathrm{M_j} & \text{if $\left|\mathrm{M_j}\right|>3.5 $} \\
            R_v & \text{otherwise}
        \end{cases}
\end{equation}
\normalsize
By incorporating two residuals, the state is updated according to \eqref{eq:MAP}, where $M$ includes $L$ and $R$. The optimal state $\bar{\textbf{x}}_{k+1}^1$ and covariance $\bar{\mathbf{P}}_{k+1}^1$ from the first stage update are then utilized in the second stage along with the gravity residual.


\subsection{Gravity Residual for Second Stage Update}
\label{subsec:second_update}
As will be shown in this paper, incorporating radar velocity measurements into \ac{LIO} effectively enhances the state estimation. Furthermore, the velocity-aware approach enables more precise local gravity estimation using the optimal state of first update. By utilizing the \ac{IMU} preintegration with estimated biases, a velocity constraint $\beta_{k+1}^{k}$ for the time interval can be derived as follows \cite{qin2018vins}: 

\vspace{-4mm}\small
\begin{eqnarray}
  \label{eq:IMU preintegration}
    \beta_{k+1}^k=\int_{t\in[r_k,r_{k+1})}{}^{I_k}\textbf{R}_{I_t}(\widehat{a}_t-\widehat{b_a}_t)dt~,
\end{eqnarray}
\normalsize
where subscript ${I_t}$ denotes the \ac{IMU} frame at index $t$. Using $\beta_{k+1}^{k}$, we can establish the relationship between \ac{IMU} acceleration measurement and estimated velocity which was updated with radar measurements:

\vspace{-4mm}\small
\begin{eqnarray}
  \label{eq:gravity relation}
    \beta_{k+1}^k={}^{I_k}\textbf{R}_{I_{k+1}}{}^{I_{k+1}}\textbf{v}_{I_{k+1}}+{}^{I_k}\textbf{g}\Delta t-{}^{I_k}\textbf{v}_{I_k}
\end{eqnarray}
\normalsize

When solving for gravity $\textbf{g}$ and expressing it in terms of the state, it can be written as follows:

\vspace{-3mm}\small
\begin{align}
  \label{eq:gravity term}
   {}^{I_k}\widehat{\textbf{g}}&=\frac{\beta^{k}_{k+1}-{}^G\bar{\textbf{R}}_{I_k}^\top{}^G\widehat{\textbf{R}}_{I_k+1}{}^G\widehat{\textbf{R}}_{I_k+1}^\top{}^G\widehat{\textbf{v}}_{I_{k+1}}+{}^G\bar{\textbf{R}}_{I_k}^\top{}^G\bar{\textbf{v}}_{I_k}}{\Delta t} \nonumber \\
   {}^G\widehat{\textbf{g}}&={}^G\bar{\textbf{R}}_{I_k}{}^{I_k}\widehat{\textbf{g}}
   =\frac{{}^G\bar{\textbf{R}}_{I_k}\beta^k_{k+1}-{}^G\widehat{\textbf{v}}_{I_{k+1}}+{}^G\bar{\textbf{v}}_{I_k}}{\Delta t}
\end{align}
\normalsize

World frame gravity ${}^G\textbf{g}$ is calculated through static initialization using \ac{IMU} acceleration. By calculating the residual between ${}^G\textbf{g}$ and the predicted gravity ${}^G\widehat{\textbf{g}}_r$ on the $\mathcal{S}^2$ manifold from ${}^G\widehat{\textbf{g}} \in \mathbb{R}^3$, compensating the deviation of roll and pitch directions is possible. Gravity residual is defined using cosine similarity, and its Jacobian is represented as follows:

\vspace{-4mm}\small
\begin{align}
  \label{eq:gravity measurement model}
    \textbf{h}_g(\textbf{x}_k,\textbf{v}_k^g)&=1-{}^G\textbf{g}^\top{}^G\widehat{\textbf{g}}_r(\textbf{x}_k,\textbf{v}_k^g) 
    =\textbf{h}_g(\widehat{\textbf{x}}_k,0)+\textbf{H}_{g}\delta\widehat{\textbf{x}}_k+n_{g} \nonumber \\
    \textbf{H}_{g}&=
    \frac{-1}{||{}^G\widehat{\textbf{g}}_r||\Delta t}\left[ (\lfloor{}^G\textbf{g}\rfloor_\times{}^G\widehat{\textbf{v}}_{I_{k+1}})^\top\quad -{}^G\textbf{g}^\top\quad\textbf{0}\quad\textbf{0}\quad\textbf{0}\right] \nonumber\\
    \textbf{r}_{g}&+\textbf{H}_{g}\delta\widehat{\textbf{x}}_k=-n_{g}\sim\mathcal{N}(0,R_g)
\end{align}
\normalsize

Since $\dim(r_g) < \dim(\mathcal{M})$, the Kalman gain is computed differently as $\textbf{K}=\mathbf{P}\textbf{H}^\top(\textbf{H}\mathbf{P}\textbf{H}^\top+\textbf{R})^{-1}$. Apart from this difference, the process remains similar to the first update.

\begin{table}[t]
\scriptsize
\caption{Effect of Velocity Measurement on Gravity Estimation Evaluated on Car Sequences from the NTU4DRadLM and Snail-Radar Datasets.}
\label{tab:velocity measurement}
\vspace{-1mm}
\centering
\resizebox{0.48\textwidth}{!}{%
\begin{tabular}{c|cc|cc|cc} \toprule
         & \multicolumn{2}{c|}{\textit{loop2}} & \multicolumn{2}{c|}{\textit{if}} & \multicolumn{2}{c}{\textit{sl}} \\\cmidrule[0.4pt](r{0.125em}){1-1} \cmidrule[0.4pt](r{0.125em}){2-7}
        & mean        & std         & mean           & std            & mean           & std            \\\cmidrule[0.4pt](r{0.125em}){1-1}\cmidrule[0.4pt](r{0.125em}){2-3} \cmidrule[0.4pt](r{0.125em}){4-5} \cmidrule[0.4pt](r{0.125em}){6-7}
\texttt{w/ vel}     & \textbf{0.035}     & \textbf{0.053}     & \textbf{0.036}        & \textbf{0.030}        & \textbf{0.164}         & \textbf{0.123}         \\
\texttt{w/o vel} & 0.037     & 0.055     & 0.043        & 0.031        & 0.167         & 0.123        \\
\bottomrule
\end{tabular}%
}

\vspace{-10mm}
\end{table}

\section{experiment}
\label{sec:experiment}

\subsection{Datasets and Evaluation Metric}

We evaluated GaRLIO on three datasets: NTU4DRadLM \cite{zhang2023ntu4dradlm}, which uses a solid-state LiDAR, and Snail-Radar \cite{huai2024snail} and $\textit{Fog-Filled hallway}$ \cite{DR-LRIO}, both employing spinning LiDARs.
NTU4DRadLM, collected using handcart and car in semi-structured environments, includes $\textit{loop*}$ sequences with significant elevation changes. Snail-Radar, collected with handheld devices, e-bikes, and SUVs in adverse weather like heavy rain, presents high dynamics in structured environments. $\textit{Fog-Filled hallway}$, gathered by drone in a geometrically uninformative area, was used only for gravity evaluation due to the lack of ground truth.

We compared GaRLIO with the three state-of-the-art LIOs: FAST-LIO2 \cite{FAST-LIO}, Point-LIO \cite{POINT-LIO}, and DLIO \cite{DLIO}. Although it would be ideal to compare our method with other radar-LiDAR-Inertial odometry approaches such as DR-LRIO \cite{DR-LRIO}, the unavailability of their code has precluded such evaluations. Consequently, we selected our comparison targets from established LIO algorithms.

\begin{figure}[!t]
    \centering
    \includegraphics[trim=0cm 0.2cm 0cm 0cm, clip, width=0.97\columnwidth]{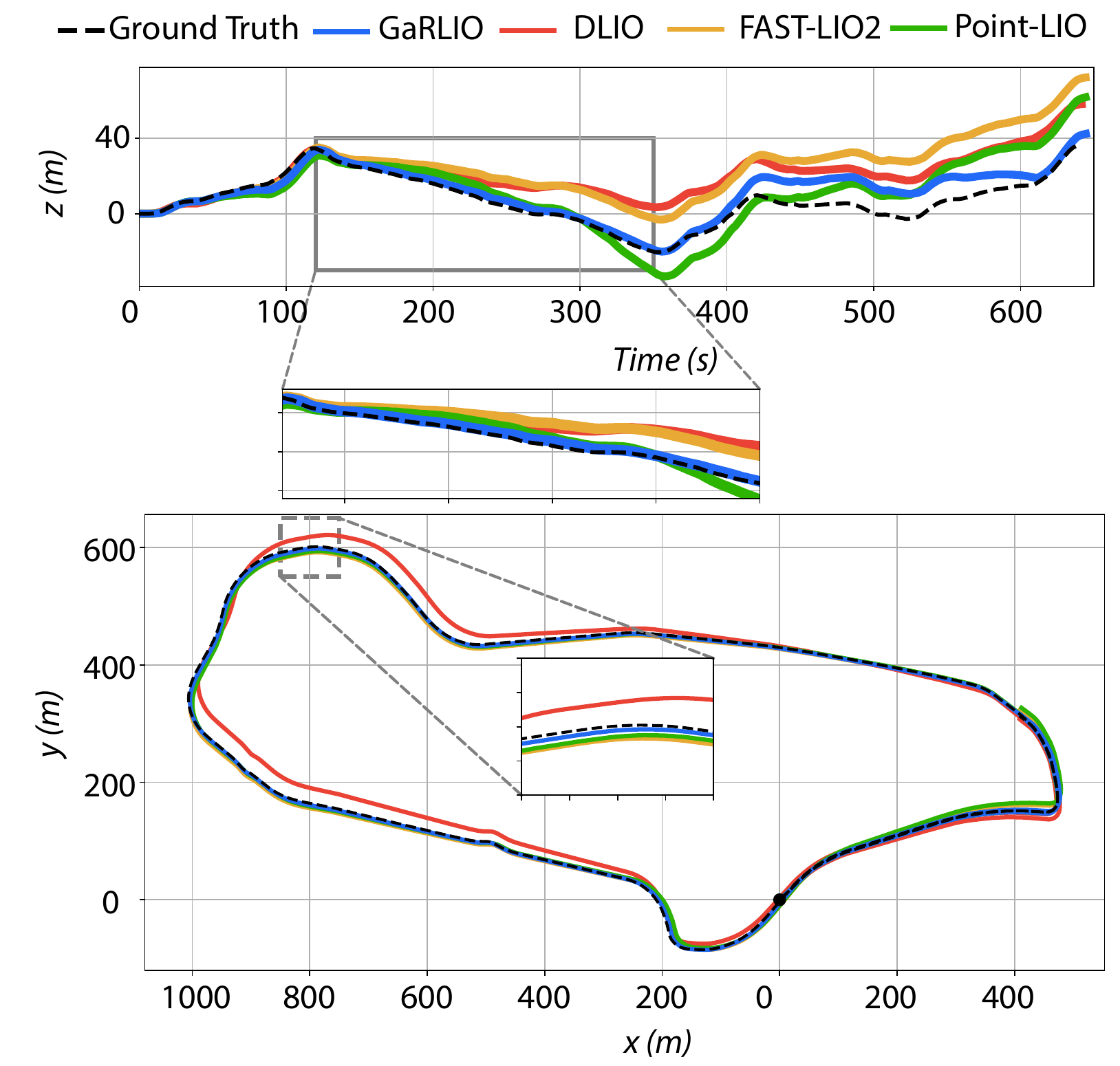}
    \caption{Estimated $z$ (top) and $xy$ (bottom) trajectory for the $\textit{loop3}$ sequence. $\textbf{Top}$ : From 120 to 350 seconds, GaRLIO accurately predicts elevation during the downhill descent, outperforming other methods. $\textbf{Bottom}$ : GaRLIO results closely match the ground truth. The black dot represents the start point. }
    \label{fig:loop3}
    \vspace{-7mm}
\end{figure}

We computed the \ac{RMSE} of the \ac{ATE} using evo \cite{grupp2017evo}, with translation errors in meters and angular errors in degrees. Results are \textbf{bold} for best, and \underline{underlined} for second best. Legends for \figref{fig:iaef_traj} and \figref{fig:sl_sequence} are the same as in \figref{fig:loop3}.

\begin{table}[t]
\scriptsize
\caption{NTU4DRadLM Dataset Evaluation}
\label{tab:NTU4DRadLM}
\vspace{-1mm}
\centering
\resizebox{0.45\textwidth}{!}{%
\begin{tabular}{cccccc} \toprule
           \multicolumn{1}{l}{}        &            & Fast LIO2 & Point LIO & DLIO   & GaRLIO
\\ \midrule
\multirow{2}{*}{\textit{cp}}         & \multicolumn{1}{|c|}{ATE$_t$} &  \textbf{0.103}    & 0.162    & 0.174 & \ul{0.107} \\
                            & \multicolumn{1}{|c|}{ATE$_r$}   &  \textbf{0.574}    & \ul{0.706}    & 1.697  & 0.750
\\ \midrule

\multirow{2}{*}{\textit{nyl}}        & \multicolumn{1}{|c|}{ATE$_t$} & 1.259     &  \textbf{0.507}    & 1.214  & \ul{0.973} \\
                            & \multicolumn{1}{|c|}{ATE$_r$}   & \ul{0.996}    &  \textbf{0.722}    & 1.688  & 1.454 \\
\midrule

\multirow{2}{*}{\textit{loop3}}      & \multicolumn{1}{|c|}{ATE$_t$} & 11.94     & 9.548     & \ul{8.332}  &  \textbf{4.271}  \\
                            & \multicolumn{1}{|c|}{ATE$_r$}   & \ul{1.714}     & 2.082     & 2.101  &  \textbf{1.496}  \\
\midrule

\multirow{2}{*}{\textit{loop2}}      & \multicolumn{1}{|c|}{ATE$_t$} & 15.81     & 20.27     & \ul{10.38}  &  \textbf{3.976}  \\
                            & \multicolumn{1}{|c|}{ATE$_r$}   & \ul{2.176}     & 4.188     & 2.525  &  \textbf{1.043}  \\ 
\bottomrule
\end{tabular}%
}
\vspace{-4mm}
\end{table}

\subsection{Evaluation on Velocity-aware Gravity Estimation}
\label{subsec:gravity test}
We performed a comparative analysis between our velocity-aware gravity estimation method and velocity-ignorant approach that utilizes double integration. The evaluation was performed across different platforms, including \ac{UGV} $(\textit{loop2}$), and drone ($\textit{Fog-Filled hallway}$). For gravity evaluation, we measured the deviation from the ground truth using the $\boxminus$ operation on $\mathcal{S}^2$ \cite{he2021kalman}. The global gravity ground truth is obtained through stationary initialization. As illustrated in \figref{fig:double integration}, our method exhibits robust gravity estimation results, maintaining minimal deviation. The disparity between velocity-ignorant and velocity-aware methods is more significant on the \ac{UGV} platform. 
\ac{UGV} introduces significant bias and noise in \ac{IMU} measurements arise from contact friction. These errors are exacerbated due to double integration, thus resulting in inaccuracy. In conclusion, our velocity-aware approach is more feasible for predicting gravity.

The effect of leveraging radar measurements on gravity estimation is depicted in \tabref{tab:velocity measurement}. Exploiting velocity updates with radar measurement consistently enhances the accuracy of estimated gravity, thus highlighting the effectiveness of radar measurement in gravity estimation.

\subsection{NTU4DRadLM Dataset}
\label{subsec:ntu}

The evaluation results for the NTU4DRadLM dataset are shown in \tabref{tab:NTU4DRadLM}. In the \textit{loop2} and \textit{loop3} sequences, where data were collected using a car, our method outperformed other LIO methods in both translation and rotation. \figref{fig:loop3} shows the qualitative results in \textit{loop3}. While other baselines show degraded performance due to steep uphill and downhill segments, GaRLIO achieves accurate estimation results in such challenging scenarios. This exhibits the robustness of our velocity-aware gravity estimation, even with significant elevation changes.
However, in $\textit{nyl}$, Point-LIO achieved the best performance.
In \textit{nyl}, the data was collected using a handcart, which causes significant vibrations.
Point-LIO effectively handles these noisy measurements by estimating acceleration and angular velocity as part of the state.
GaRLIO is not specifically designed for such conditions, although it achieves comparable performance.

%

\begin{table}[t]
\scriptsize
\caption{Snail Radar Dataset Evaluation}
\label{tab:Snail-Radar}
\vspace{-1mm}
\centering
\resizebox{0.45\textwidth}{!}{%
\begin{tabular}{cccccc} \toprule
           \multicolumn{1}{l}{}        &            & Fast LIO2 & Point LIO & DLIO   & GaRLIO \\ \midrule
\multirow{2}{*}{ \textit{sl}} & \multicolumn{1}{|c|}{ATE$_t$} & 14.34     & 10.98     & \ul{5.054}  &  \textbf{1.059}  \\
                            & \multicolumn{1}{|c|}{ATE$_r$}   & 3.318     & 3.321     & \ul{2.718}  &  \textbf{1.170}  \\ \midrule
\multirow{2}{*}{\textit{if}} & \multicolumn{1}{|c|}{ATE$_t$} & 5.747     & 9.018     &  \textbf{0.308} & \ul{0.450} \\
                            & \multicolumn{1}{|c|}{ATE$_r$}   & 1.479     & 3.291     & \ul{0.730} &  \textbf{0.591} \\ \midrule
\multirow{2}{*}{\textit{iaf}} & \multicolumn{1}{|c|}{ATE$_t$} & 24.99     & 73.19     & \ul{6.192}  &  \textbf{4.672}  \\
                            & \multicolumn{1}{|c|}{ATE$_r$}   & 2.826     & 9.542     & \ul{1.560}  &  \textbf{1.539}  \\ \midrule
\multirow{2}{*}{\textit{iaef}} & \multicolumn{1}{|c|}{ATE$_t$} & 30.26     & 90.81     & \ul{15.86}  &  \textbf{6.450}  \\
                            & \multicolumn{1}{|c|}{ATE$_r$}   & 3.317     & 9.004     & \ul{1.436}  &  \textbf{1.397}  \\
\bottomrule
\end{tabular}%
}
\vspace{-3mm}
\end{table}
\begin{figure}[!t]
    \centering
    \includegraphics[trim=0cm 0.4cm 0cm 0cm, clip,width=0.97\columnwidth]{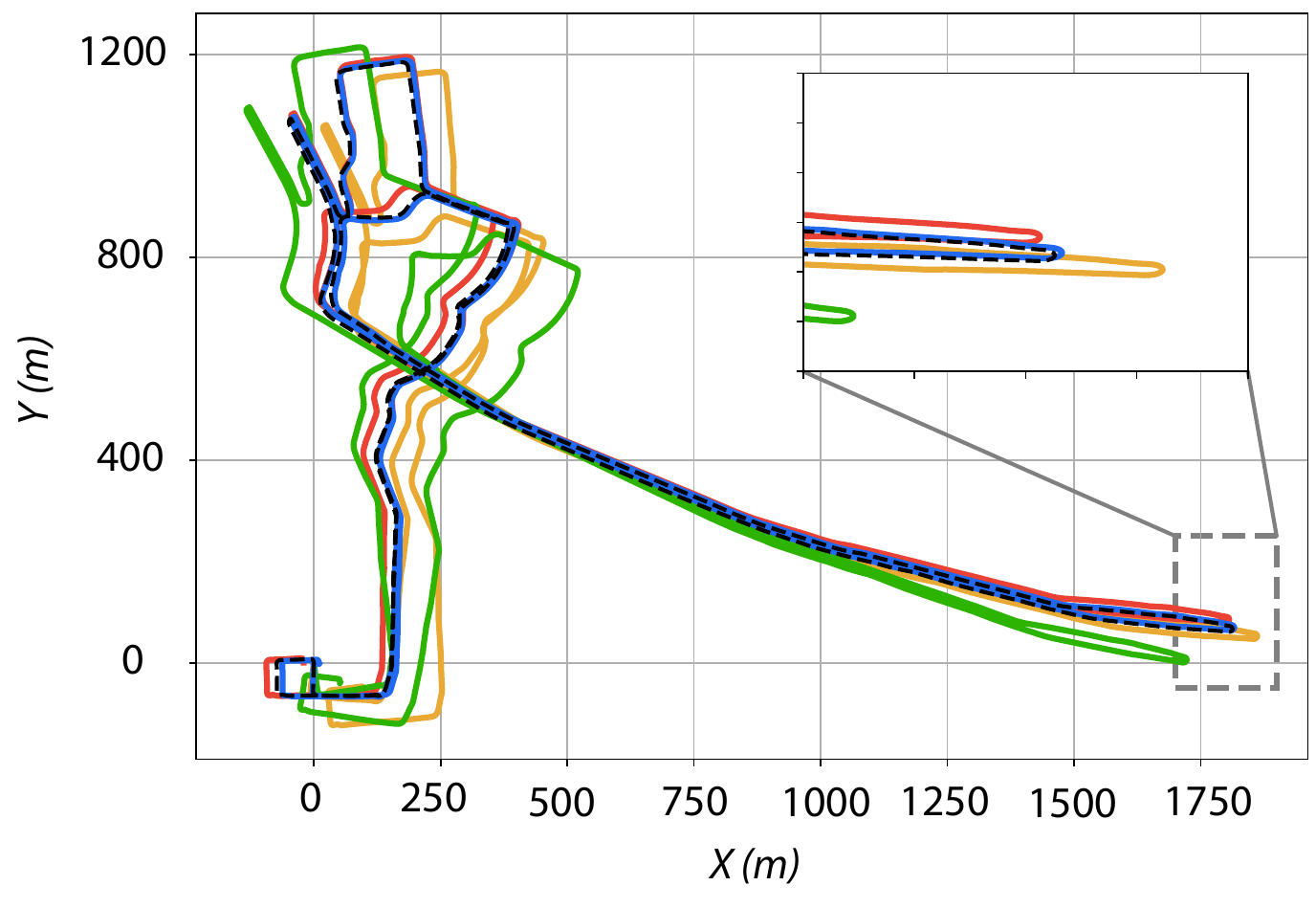}
    \caption{Estimated trajectory for the $\textit{iaef}$ sequence and GaRLIO demonstrated superior performance over such a long sequence. 
 }
    \label{fig:iaef_traj}
    \vspace{-7mm}
\end{figure}

\subsection{Snail-Radar dataset}
\label{subsec:snail}

\begin{figure}[!t]
    \centering
    \includegraphics[trim=0cm 0.3cm 0cm 0cm, clip, width=1\columnwidth]{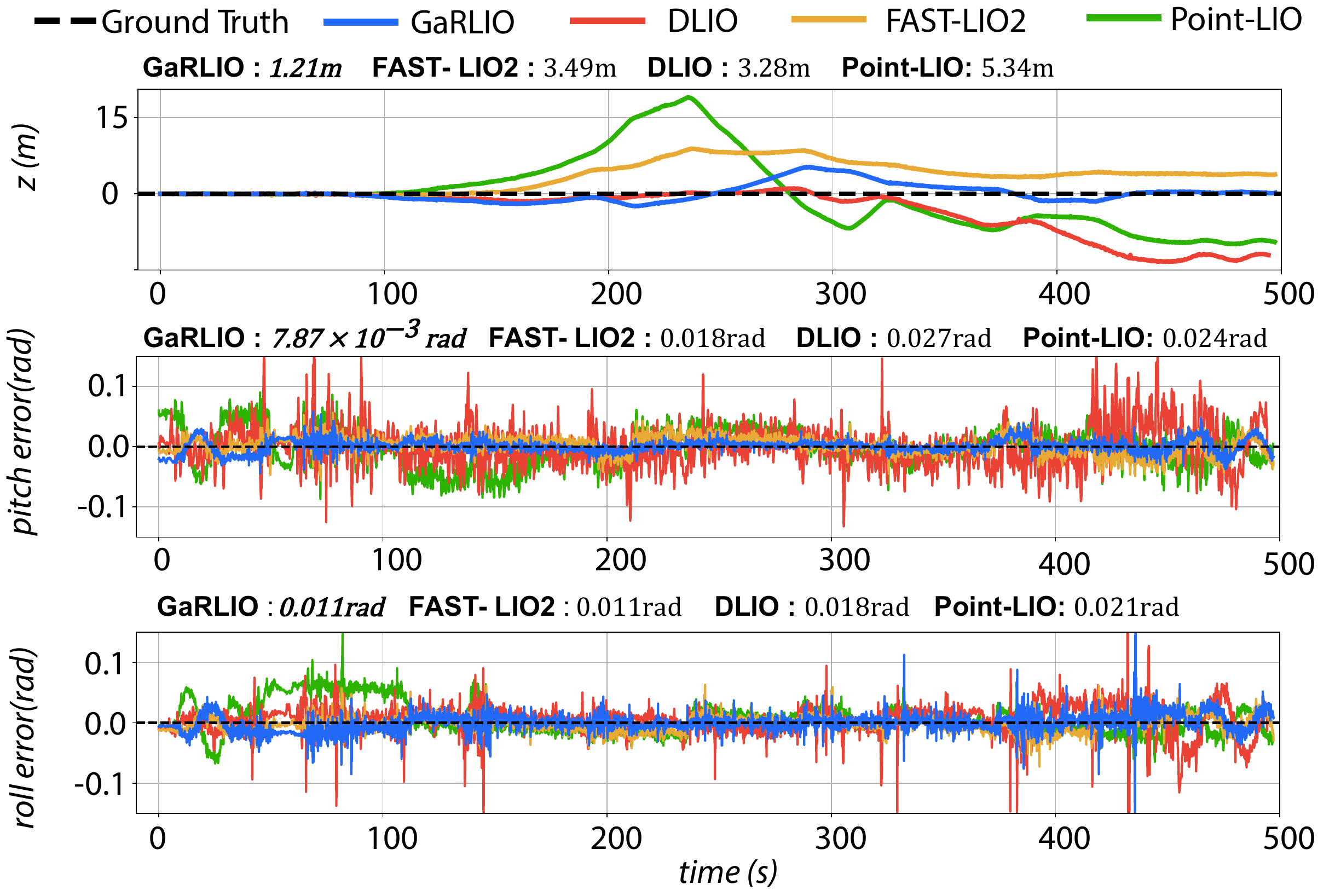} 
    \caption{$z$, pitch, and roll error in $\textit{sl}$. The RMSE values for $z$, pitch, and roll are above each plot. Our method shows the lowest error along the baselines.
 }
    \label{fig:sl_sequence}
    \vspace{-3mm}
\end{figure}
\begin{figure}[]
    \centering
    \includegraphics[width=0.95\columnwidth]{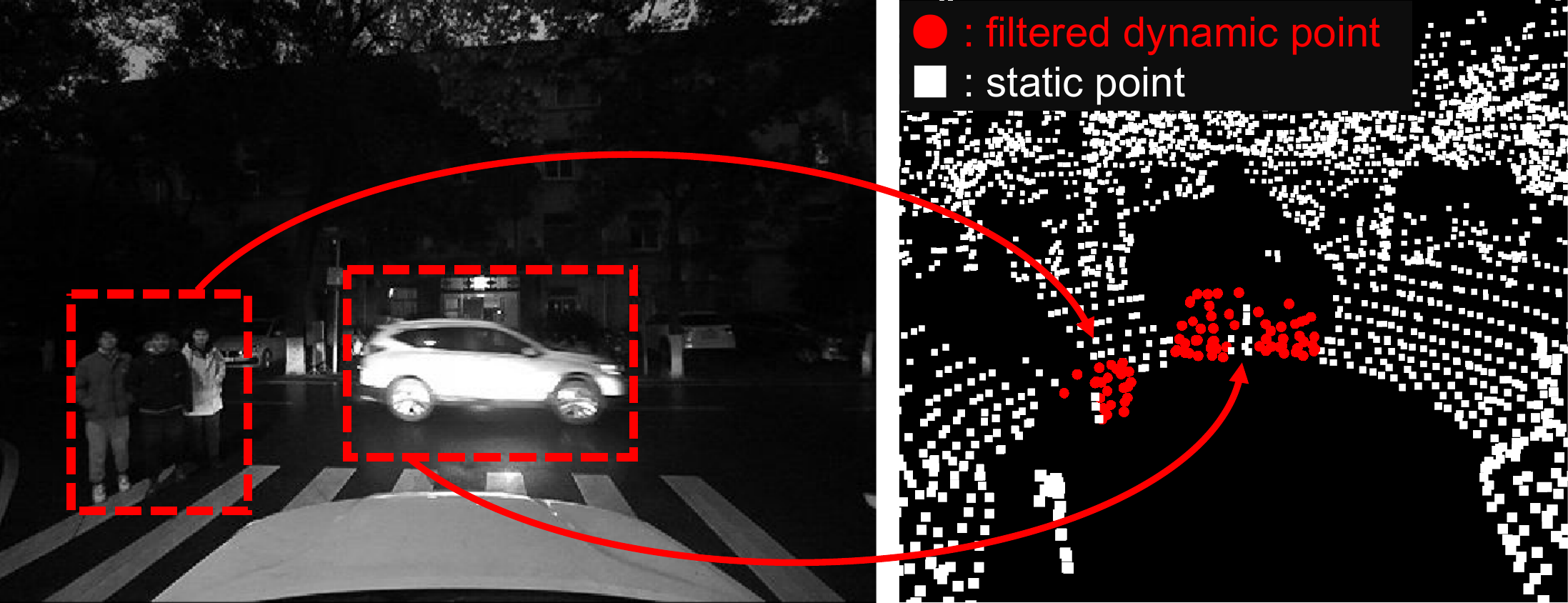}
    \caption{Result of Dynamic removal module in Snail-Radar. The red points in the right LiDAR point cloud represent dynamic points that were removed using radar-based dynamic point.}
    \label{fig:dynamic}
    \vspace{-7mm}
\end{figure}

As shown in \tabref{tab:Snail-Radar} and \figref{fig:iaef_traj}, our method outperforms other LIO algorithms throughout the experiments. 
Specifically, in the $\textit{sl}$ sequence, other baselines exhibit significant errors in the $z$-axis due to the vigorous dynamic movements of the e-bike as shown in \figref{fig:sl_sequence}. Additionally, the Snail-Radar dataset contains various dynamic objects, such as vehicles and pedestrians, which affected the state predictions of FAST-LIO2 and Point-LIO, leading to degraded RMSE performance.
Conversely, GaRLIO effectively mitigates drift in the roll and pitch directions through the velocity-aware gravity residuals, resulting in robust elevation estimation. Furthermore, GaRLIO successfully removes dynamic objects within the LiDAR by leveraging radar, as shown \figref{fig:dynamic}, resulting in more reliable performance compared to other methods.
DLIO shows second-best performance across most sequences attributed to Geometric Observer and a continuous-time method. However, the absence of a dynamic handling mechanism and an update module based on velocity information reveals limitations along the experiments.


\subsection{Ablation Study}
\label{subsec:ablation}

To evaluate the effect of each residual and dynamic removal module, we conducted experiments using various platforms, including a handcart (\textit{cp}, \textit{nyl}), a car (\textit{loop2}, \textit{loop3}), and an e-bike (\textit{sl}). We created five variants by disabling two of the residuals ($\texttt{g\&v}$) and Dynamic removal module ($\texttt{d}$): $\texttt{w/o g\&v\&d}$, $\texttt{w/o g\&v}$, $\texttt{w/o v}$, $\texttt{w/o g}$, $\texttt{FULL}$, evaluated in terms of \ac{ATE}.

\subsubsection{Effect of Velocity Residual}
As depicted in \tabref{tab:Ablation}, incorporating velocity residuals significantly enhances performance. This result corroborates the discussion in \secref{subsec:gravity test}, demonstrating that incorporating velocity measurements facilitates more precise gravity estimation, thereby improving odometry estimation accuracy. 
Furthermore, the \textit{loop3} and \textit{loop2} sequences,
which were collected using a car with higher speeds,
demonstrate substantial improvements in state estimation through the incorporation of velocity residuals compared to \textit{cp}, \textit{nyl}, and \textit{sl}, which are the slower sequences. This improvement is attributed to the inherent limitations of existing LIO methods that depend on kinematic models for velocity estimation, which are less accurate than our method at higher speeds.

\begin{figure}[]
    \centering
    \includegraphics[width=0.95\columnwidth]{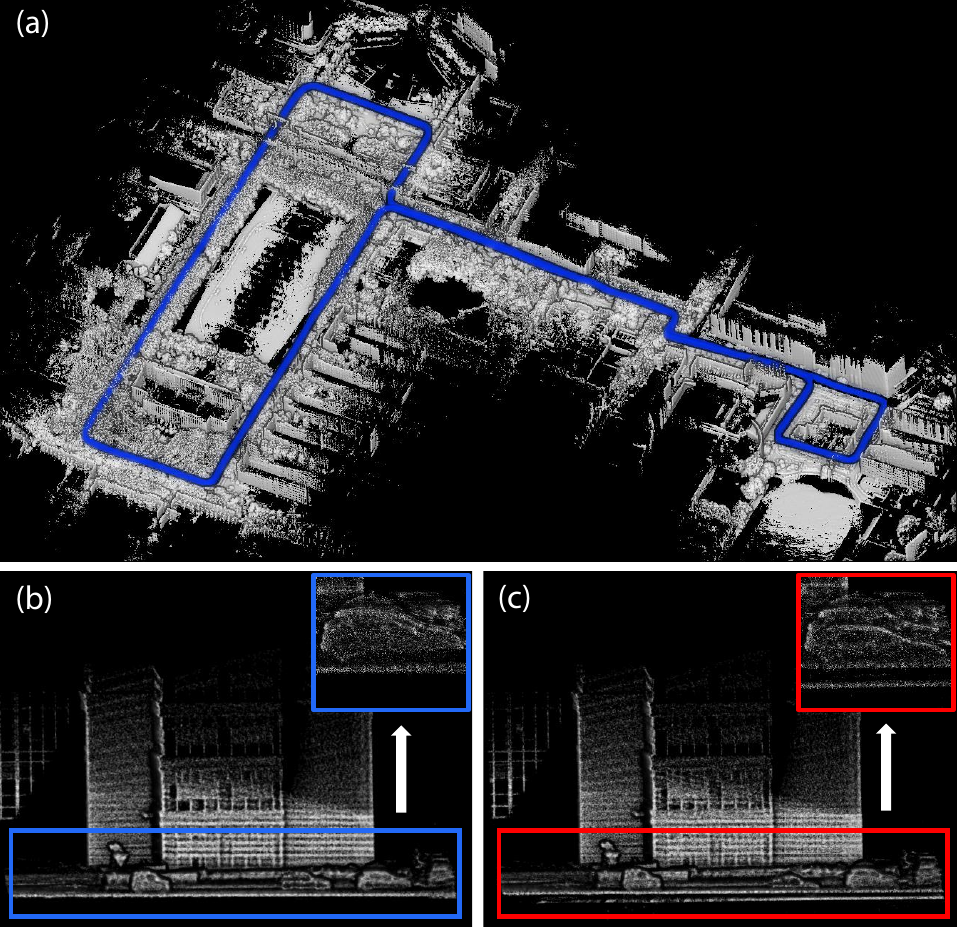}
    \caption{(a) Mapping result of the $\textit{if}$ sequence generated using GaRLIO. Our method shows accurate alignment when revisiting previously traversed areas. At the revisited starting point: (b) With gravity residuals, ground regions are accurately aligned; (c) Without gravity residuals, vertical drift becomes apparent.}
    \label{fig:gravity}
    \vspace{-2mm}
\end{figure}

\begin{table}[t]
\scriptsize
\caption{Effect on \ac{ATE} of Dynamic Removal Module and Each Residual}
\label{tab:Ablation}
\vspace{-1mm}
\centering
\resizebox{0.45\textwidth}{!}{%
\begin{tabular}{c|ccccc} \toprule
                & \textit{cp}     & \textit{loop3}  & \textit{loop2}  & \textit{nyl}    & \textit{sl}    \\ \cmidrule[0.4pt](r{0.125em}){1-1} \cmidrule[0.4pt](r{0.125em}){2-6}
\texttt{w/o g\&v\&d} & 0.101 & 12.44  & 9.163  & 1.281  & 1.889 \\
\texttt{w/o g\&v} & 0.115 & 10.83  & 8.573  & 1.280  & 1.467 \\
\texttt{w/o g}  & 0.107 & 9.952  & 5.985  & 1.135  & 1.129 \\
\texttt{w/o v}   & \textbf{0.098} & 11.28 & 11.17 & 0.991 & 1.194 \\ \cmidrule[0.4pt](r{0.125em}){1-1} \cmidrule[0.4pt](r{0.125em}){2-6}
\texttt{FULL}         & 0.107 & \textbf{4.271}  & \textbf{3.976}  & \textbf{0.973} & \textbf{1.059}  \\                  
\bottomrule
\end{tabular}%
}
\vspace{-6mm}
\end{table}

\subsubsection{Effect of Gravity Residual}
When using gravity residuals without the addition of velocity measurements, a decline in performance was observed in $\textit{loop3}$ and $\textit{loop2}$, as indicated by the comparison between $\texttt{w/o g\&v}$ and $\texttt{w/o v}$. 
As shown in \tabref{tab:velocity measurement}, GaRLIO relies on the accuracy of the velocity for precise gravity estimation, which likely leads to performance degradation in the absence of velocity measurements. 
Accurate gravity estimation through velocity updates significantly improves odometry performance, as demonstrated by the comparison between $\texttt{w/o g}$ and $\texttt{FULL}$. These enhancements are particularly noticeable in elevation estimation, as seen in \figref{fig:gravity}.

\subsubsection{Effect of Dynamic Removal Module}
The impact of the dynamic removal module can also be observed in \tabref{tab:Ablation}.
Removing dynamic objects using radar point clouds improves performance in most sequences, with particularly notable enhancements in those containing numerous dynamic objects, such as \textit{loop3}, \textit{loop2}, and \textit{sl}.
These results demonstrate that the proposed algorithm is robust in environments with a high density of dynamic points.

\section{Conclusion}
\label{sec:conclusion}

In this paper, we introduce GaRLIO, a gravity-enhanced radar-LiDAR-inertial odometry that provides a novel gravity estimation method that utilizes radar Doppler measurements.
Differing from the velocity-ignorant approaches, our method ensures robust gravity estimation along the various platforms.
Furthermore, the fusion of radar with LIO facilitates the dynamic removal within LiDAR point clouds.
We validated its performance in public datasets, demonstrating its robustness even in challenging scenarios such as downhill and dynamic object-rich conditions.
Remarkably, our approach represents superior improvements in mitigating vertical drift. GaRLIO, the first method to combine radar and gravity,  is anticipated to establish new research directions for advancing robust SLAM systems based on \ac{UGV}.

\newpage

\bibliographystyle{IEEEtranN} 
\bibliography{string-short,references}

\end{document}